\documentclass{article}

\usepackage{arxiv}

\usepackage[utf8]{inputenc} 
\usepackage[T1]{fontenc}    
\usepackage{hyperref}       
\usepackage{url}            
\usepackage{booktabs}       
\usepackage{amsmath,amssymb,amsfonts}       
\usepackage{nicefrac}       
\usepackage{microtype}      
\usepackage{lipsum}		
\usepackage{graphicx}
\usepackage{natbib}
\usepackage{doi}

\usepackage{mathtools}
\usepackage{verbatim}

\usepackage{algorithm}
\usepackage{algpseudocode}

\DeclareMathOperator*{\argmax}{argmax}
\newtheorem{theorem}{Theorem}

\NewDocumentCommand{\camp}{}
{
  \unskip\textsc{C\small a\small m\small p}\unskip
}

\makeatletter
\newcommand*{\rom}[1]{\expandafter\@slowromancap\romannumeral #1@}
\makeatother

\title{Contextual Augmented Multi-Model Programming (CAMP): A Local-Cloud Copilot Solution}

\author{ \href{https://orcid.org/0000-0002-5183-1248}{\includegraphics[scale=0.06]{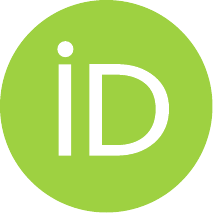}\hspace{1mm}Yuchen Wang} \\
	Nanyang Technological University\\
	Singapore \\
	\texttt{yuchen011@e.ntu.edu.sg} \\
	\And
	\href{https://orcid.org/0009-0004-0779-7179}{\includegraphics[scale=0.06]{orcid.pdf}\hspace{1mm}Shangxin Guo} \\
	City University of Hong Kong\\
	Hong Kong \\
	\texttt{sxguo2-c@my.cityu.edu.hk} \\
    \And
	\href{https://orcid.org/0000-0002-6624-9752}{\includegraphics[scale=0.06]{orcid.pdf}\hspace{1mm}Chee Wei Tan} \\
	Nanyang Technological University\\
	Singapore \\
	\texttt{cheewei.tan@ntu.edu.sg} \\
}

\date{}

\renewcommand{\shorttitle}{\textit{arXiv} Template}

\hypersetup{
pdftitle={A template for the arxiv style},
pdfsubject={q-bio.NC, q-bio.QM},
pdfauthor={David S.~Hippocampus, Elias D.~Striatum},
pdfkeywords={First keyword, Second keyword, More},
}

\begin{document}
\maketitle

\begin{abstract}
The rapid advancement of cloud-based Large Language Models (LLMs) has revolutionized AI-assisted programming, but their integration into local development environments faces trade-offs between performance and cost. Cloud LLMs deliver superior generative power but incur high computational costs and latency, whereas local models offer faster, context-aware retrieval but are limited in scope. To address this, we propose \camp{}, a multi-model copilot solution that leverages context-based Retrieval Augmented Generation (RAG) to enhance LLM performance through dynamic context retrieval from local codebases which optimizes context-aware prompt construction. Experimental results show \camp{} achieves a 12.5\% improvement over context-less generation and 6.3\% over the basic RAG approach. We demonstrate the methodology through the development of ``Copilot for Xcode," which supports generative programming tasks including code completion, error detection, and documentation. The tool gained widespread adoption and was subsequently integrated into GitHub Copilot, highlighting \camp{}'s impact on AI-assisted programming and its potential to transform future software development workflows.
\end{abstract}


\section{Introduction}
Dijkstra, in his work \cite{dijkstra,dijkstra1972humble}, proposed computer-assisted programming, emphasizing the breakdown of complex programs into smaller, deliberate decisions to prevent bugs and improve understanding—a vision now realized through AI-assisted programming powered by large language models (LLMs) \cite{mozannar2022reading,wong2023natural}. These models, driven by advancements in natural language processing (NLP), automate code generation and enable interactive software development. Similarly, Sammet, in her 1966 work \cite{Sammet1966}, explored the use of English as a programming language, highlighting its potential to make programming more accessible and intuitive. Addressing runtime program modification and incorporating multi-modal feedback are crucial for enhancing problem composition quality. Today, developers leverage AI-driven capabilities to enhance efficiency and productivity in software development.

One of the earliest AI-assisted programming tools, the {\it MIT Programmer's Apprentice}, simulated a skilled junior programmer, leveraging NLP to analyze and understand programming patterns \cite{MITapprentice82,MITapprentice88}. It introduced concepts such as code generation \cite{MITcodegeneration} and an early form of ``prompt engineering" \cite{MITprompt}, recognizing programming as a process of abstraction and simplification \cite{MITsimplify}. Advances in AI-assisted programming are now made by leveraging prompt engineering, in-context learning, and crowdsourcing of human feedback alongside large codebases for unsupervised learning \cite{wong2023natural}.

\begin{figure}[!t]
\centering
\includegraphics[width=0.8\linewidth]{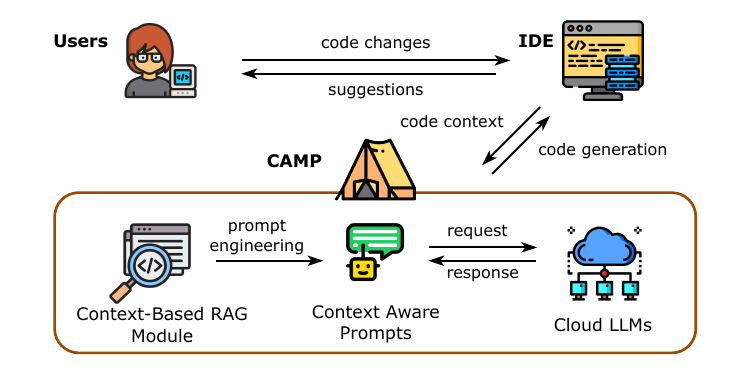}
\caption{An overview of \camp{}, the AI-assisted programming solution that empowers cloud LLM with local code context information retrieved by context-based RAG. }\label{fig:overview}
\vspace{-5mm}
\end{figure}

Cloud-based tools leveraging LLMs, such as Codeium \cite{codeium}, GitHub Copilot \cite{GitHubCopilot, GitHubCopilot_2022}, OpenAI ChatGPT \cite{gpt42023}, Amazon CodeWhisperer \cite{amazon}, and Meta's Code Llama \cite{meta_llama}, provide users with access to LLM services through dedicated APIs on demand. These tools can be integrated into existing systems or implemented via Software-as-a-Service (SaaS) web interfaces, acting as virtual service entities to meet objectives and reduce costs for programmers \cite{zheng2015bid}. The reach and high demand for these LLM-based tools reflect the growing need for advanced NLP capabilities in software development, resonating with Dijkstra's vision of a paradigm shift, where the challenge lies not just in executing programs but in their creation and maintenance \cite{dijkstra1972humble,dijkstra}. Cloud-based LLMs, while offering substantial generative power, come with high computational costs and latency, and they face challenges in integrating smoothly within constrained development environments.

The advent of RAG has further revolutionized AI-assisted programming \cite{rag_lewis}. By combining the strengths of pre-trained LLMs and information retrieval techniques, RAG models can retrieve relevant documents from a large corpus and use them to condition the generation process of the language model. This approach inspires us to use a RAG-based local model to enhance prompt generation for LLMs.

This paper presents \camp{}, a multi-model copilot programming solution that leverages local code context retrieval and cloud LLMs to optimize context-aware code generation. As shown in Figure~\ref{fig:overview}, \camp{} integrates cloud LLMs into local development environments, employing a RAG module that dynamically learns from code context to optimize prompt construction. This methodology is implemented in {\it Copilot for Xcode},\footnote{\tt \url{https://github.com/intitni/CopilotForXcode}} a tool providing automatic code completion, error detection, and documentation, synchronized with user interactions and codebase updates. The project was open-sourced and later integrated into GitHub Copilot for Xcode \cite{tan2023copilot, github_xcode_2024}.

The key contributions of this study include:
\begin{itemize}
    \item{We proposed \camp{}, a multi-model framework using context-based RAG to enhance AI-assisted programming, achieving a 12.5\% improvement over context-less generation and 6.3\% over baseline RAG models on the code generation benchmark.}
    \item{We mathematically formalize the context-based RAG problem and propose generalizable algorithms to solve the resulting optimization problem.}
    \item{We develop and deploy {\it Copilot for Xcode}, an implementation of \camp{}, which achieves widespread adoption by the developer community and integration into GitHub Copilot.}
\end{itemize}

\section{Related Works}
\subsection{Software Naturalness Hypothesis}
The software naturalness hypothesis suggests that programming languages should mimic the patterns found in natural language processing \cite{hindle2012naturalness}. This concept is supported by early n-gram models for code completion, highlighting software's repetitive and predictable nature. This conceptualization of modeling codes through statistical language models underpins our approach, where we optimize prompt engineering through fine-tuning hyperparameters, setting the stage for more intuitive AI-assisted programming solutions, as detailed in Section \ref{sec:rag}.

\subsection{Language Models and AI-assisted Programming}
Since the introduction of the transformer architecture \cite{vaswani2017attention}, LLMs trained on extensive datasets have excelled in code-related tasks, contributing significantly to Big Code analysis \cite{vechev2016programming}. Models such as T5 \cite{raffel2020exploring}, BERT \cite{devlin2018bert}, GPT-4 \cite{gpt42023}, and Palm 2 \cite{anil2023palm} exhibit remarkable capabilities in understanding and generating text, thereby enhancing software development processes. AI-assisted programming leverages these models to automate tasks like code generation \cite{waldinger1969prow, wong2024aligning}, completion \cite{robbes2008program}, and translation \cite{acharya2007mining}. Tools like GitHub Copilot \cite{GitHubCopilot, GitHubCopilot_2022}, Codeium \cite{codeium}, and ChatGPT \cite{gpt42023} are widely used for generative coding tasks. Beyond code generation, LLMs also contribute to software security by enabling codebase analysis, bug detection, automated fixes, and test generation \cite{copilot_for_testing}. However, the full integration of LLMs into IDEs like Xcode remains constrained by computational costs and restricted access, leaving a gap that motivates our work to fully harness the capabilities of these models.

\subsection{Retrieval Augmented Generation (RAG)}
RAG represents a recent advancement in NLP by integrating pre-trained language models with information retrieval techniques. This approach retrieves relevant documents from large corpora to enhance the language model's generation process \cite{rag_lewis, rag_izacard}. RAG's potential extends to programming by enhancing code generation through the retrieval of pertinent code snippets from extensive source code repositories. This insight informs our work, where we leverage a code context-based local model to collaborate with cloud LLMs through context-aware prompt engineering.

\subsection{Constraints of Local Integrated Development Environments (IDEs)}
IDEs such as Xcode \cite{xcode} offer essential tools for writing, debugging, and testing software. However, integrating AI-assisted programming with LLMs into these environments presents significant challenges, arising from high computational demands \cite{ide_8}, network latency \cite{ide_9}, and limited access \cite{ide_10_apple}, which collectively constrain the capabilities of LLM-driven code generation. Addressing these limitations necessitates a solution that effectively bridges the gap between the contextual information available in local development environments and the generative power of cloud-based LLMs, which motivates our proposed multi-model solution. Furthermore, Xcode serves as a strategic starting point for implementing our methodology, with its successful application potentially extending easily to other platforms with fewer constraints.

\begin{figure*}[t]
\centering
\includegraphics[width=0.8\linewidth]{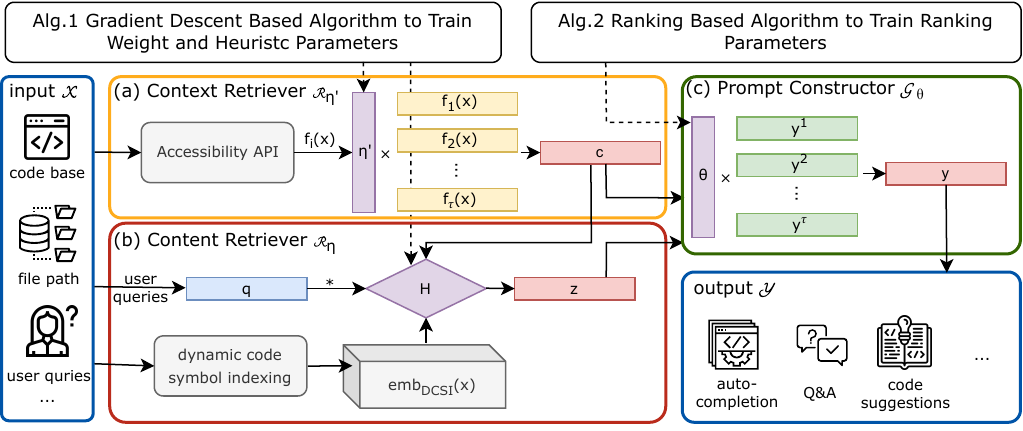}
\caption{Overview of the RAG module. (a) Context retriever $\mathcal{R}_{\eta'}$ that retrieves contextual information from the local development environment. (b) Content retriever $\mathcal{R}_\eta$ that searches for the most relevant information from local content. (c) Prompt constructor $\mathcal{G}_\theta$ that creates context-aware prompts.} \label{fig:rag_model_framework}
\end{figure*}

\section{Problem Formulation}\label{math}
In this section, we mathematically formulate the language model for AI-assisted programming with context-based RAG, establishing key problem metrics and demonstrating its feasibility.

A ``programming copilot" can be represented as a language model with inputs such as user commands, existing code, and past tokens, and outputs such as automated code completions, suggestions, and query responses. Focusing on context-based content generation, the proposed generator is formulated as a language model that takes \textit{context} as input and produces \textit{retrieved content} as output.

We start with the maximum entropy language model with the following form \cite{max_entropy}:
\begin{equation}\label{basic_model_2}
    p(w|h) = \frac{\exp{(\psi(w)^TA\phi(h))}}{\sum_{w'}{\exp{(\psi(w')^TA\phi(h))}}}
\end{equation}
where $w$ is the generated word given history $h$; $\psi(\cdot) \in \mathbb{R}^{d_{\psi}}$ and $\phi(\cdot)\in \mathbb{R}^{d_{\phi}}$ are individual embeddings of the word and history, and $A$ represents the model's parameters attached to the extracted features.

A typical RAG model uses the input sequence to retrieve relevant content (also called ``document") and then uses both the input and the document to generate the output sequence \cite{rag_lewis}. The retriever $p_\eta(z|x)$ computes the probability distribution of the top documents over the database, given input $x$; the generator $p_\theta(y_i|x, z, y_{1:i-1})$ then generates token $y_i$ based on the original input $x$ and the retrieved document $z$. The model is end-to-end formulated as
\begin{align}
\begin{split}
    p_{\mathrm{RAG}} & = 
    \sum_{z \in \mathrm{top-K}(p(\cdot|x))}{p_\eta(z|x)p_\theta(y|x, z}) \\
         & = \sum_{z \in \mathrm{top-K}(p(\cdot|x))}{p_\eta(z|x)\prod_{i}^{N}{p_\theta(y_i|x, z, y_{1:i-1})}}.
\end{split}
\end{align}

Our proposed context-based RAG module utilizes local code context $c$ to enhance the content-retrieving procedure, which yields
\begin{align}
\begin{split}
    p_{\mathrm{RAG}} 
         & = \sum_{z \in \mathrm{top-K}(p(\cdot|x))}{p_{\eta'}(c|x)p_{\eta}(z|x, c)\prod_{i}^{N}{p_\theta(y_i|x, z, y_{1:i-1})}} \label{content_rag_1}.
\end{split}
\end{align}

From (\ref{content_rag_1}), the problem to solve can then be broken down to the modeling and optimization of individual sub-models $p_\cdot(\cdot|\cdot)$, including the context retrieval model $p_{\eta'}(c|x)$, the content retrieval model $p_{\eta}(z|x, c)$, and the prompt generation model $p_\theta(y_i|x, z, y_{1:i-1})$. 

To show the problem is feasible, has a global optimal solution, and can be solved iteratively, we take the content retrieval model $p_{\eta}$ as a typical example and cater (\ref{basic_model_2}) to our use case as
\begin{align} \label{h=usv}
\begin{split}
    p_{\eta}(z|x, c) & = \frac{\exp{(\psi(z)^TH\phi(x, c))}}{\sum_{z'}{\exp{(\psi(z')^TH\phi(x, c))}}} \\
    & = 
    \frac{\exp{(\psi(z)^TU\Sigma V^T\phi(x, c))}}{\sum_{z'}{\exp{(\psi(z')^TU\Sigma V^T\phi(x, c))}}} \\
    & =
    \frac{\exp{(\hat\psi(z)\Sigma \hat\phi(x, c))}}{\sum_{z'}{\exp{(\hat\psi(z')\Sigma \hat\phi(x, c))}}}
    \end{split}
\end{align}
where $\phi(\cdot)\in \mathbb{R}^{d_{\phi}}$ extracts feature embeddings from both the original input and the context and $H \in \mathbb{R}^{d_{\psi} \times d_{\phi}}$ represents the heuristic matrix that determines the ranking of documents in the content search and is applicable for the singular value decomposition (SVD) of $H = U\Sigma V^T$. We can then claim the model defined by (\ref{h=usv}) to be in a continuous space and both $\hat\psi(z)$ and $\hat\phi(x, c)$ are continuous embeddings.

This can be formulated as a convex optimization problem
\begin{equation}\label{opt1}
    \min_{H}-\mathcal{L}(\mathcal{X}, \mathcal{Y}, H) + \mathcal{R}(H)
\end{equation}
where $\mathcal{L}(\mathcal{X}, \mathcal{Y}, H) = \frac{1}{N}\sum_{i=1}^N\log{P(y_i|x_i, H)}$ is the target function and $\mathcal{R}(H)$ is the regularization term. This can be solved using gradient descent algorithms \cite{proximal_gradient_algo}.

\section{Methodology}
We refer to the software naturalness hypothesis to give the mathematical definition of the research problems: compute a function $\mathcal{F}$ over the local development environment $x$, where the proposed model $\mathcal{M}$, with fine-tunable parameters $\gamma$, provides prompts for LLMs to obtain real-time ``suggestions" $s$ as
\begin{equation*}
    \mathcal{F}_\mathcal{M_\gamma}(s|x): \mathcal{X} \rightarrow \mathcal{S} 
\end{equation*}
where $\mathcal{X}$, represents the domain of the input information, including environment-related information (e.g. source code, current repository, and editor status) and user-related information (e.g. detected user actions and requests); $\mathcal{S}$ represents the domain of the output provided by $\mathcal{M}$, including auto-completed code, error warning messages, answers to explicit requests in the chat panel, and so on. In the following sections, we present our solution with details of $\mathcal{M}$ and propose algorithms to finetune its parameters $\gamma$.

\subsection{Context-Based RAG} \label{sec:rag}
As defined in Section \ref{math}, our proposed RAG module consists of three major components: (\rom{1}) a context retriever $\mathcal{R}_{\eta'}(c|x)$ that captures contextual information from the local development environment, (\rom{2}) a content retriever $\mathcal{R}_{\eta}(z|x,c)$ that generates relevant content given the current context and the original input, and (\rom{3}) a prompt constructor $\mathcal{G}_\theta(y_i|x, c, z, y_{1:i-1})$ that creates prompts to assist LLMs from the retrieved information and user queries.

Figure~\ref{fig:rag_model_framework} presents a detailed illustration of the system workflow. Given the local development environment at a certain timestamp $t$, the contextual information $c$ is first obtained and utilized for the retrieval of the top-ranked relevant content information $z$. Both the context $c$ and content $z$ are then utilized in prompt construction for LLMs requests. As the local development environment evolves with $t$, this workflow synchronizes with user actions and codebase changes, providing on-demand functionalities.

In the following subsections, we detail the three components sequentially.

\begin{table}[t]
    \centering
    \caption{Major Components of a Constructed Prompt. Components are ranked in decreasing priorities computed with Algorithm \ref{alg2}.}
    \begin{tabular}{cc}
        \toprule
        \textbf{Component} & \textbf{Priority} \\
        \midrule
        Context System Prompt (by $\mathcal{R}_{\eta'}(c|x)$) & High \\
        Retrieved Content (by $\mathcal{R}_{\eta}(z|x,c)$) & High \\
        New Message & High \\
        Message History & Medium \\
        System Prompt & Low \\
        \bottomrule
    \end{tabular}
    \label{tab:components}
    \vspace{-5mm}
\end{table}

\subsubsection{Context Retriever}
The context retriever obtains contextual information from the local development environment that maximizes the insights brought to the next step. Many factors in the input environment might be considered, including the user's point of view, the current file opened, and highlighted code snippets, though we can not afford to cover all possible aspects without ``over-sparsing" the feature vectors or causing computational burdens. We define $\tau_{c}$ to be the upper limit of the contextual entries to include. We then have
\begin{align} \label{eqn_c}
\begin{split}
    \mathcal{R}_{\eta'}(x) & = \mathrm{agg}([\eta'_0c_0, \eta'_1c_1, \dots, \eta'_{\tau_{c}}c_{\tau_{c}}]) \\
    & = \mathrm{agg}([\eta'_0f_0(x_0), \eta'_1f_1(x_1), \dots, \eta'_{\tau_{c}}f_{\tau_{c}}(x_{\tau_{c}})]) \\
    & = \mathrm{agg}(\eta'\cdot f(x))
\end{split}
\end{align}
where we abuse the annotation $f_i(\cdot)$ to represent the detailed data processing for each contextual entry and $\mathrm{agg}$ to represent the aggregation method. We normalize by setting $\Sigma\eta'_i = 1$ and assign a larger value to $\eta'_i$ to increase the influence of the corresponding $c_i$. For null entries, where the number of selected components is below the limit $\tau_c$, we set $\eta'_i = 0$.

We eventually select ``cursor position", ``absolute repository path", ``cached build artifacts", and ``index information" as our sources of contextual information based on trials and errors. The weight parameters, $\eta'$, are fine-tuned using the algorithm outlined in Section \ref{parameter_tuning}. With the assumption that the relative importance of different factors in the local development environment remains stable, we can obtain a fixed set of optimal $\eta'$ values over time and across data $(\mathcal{X}, \mathcal{Y})$.

\subsubsection{Content Retriever}
The objective of the content retriever is to deliver highly relevant content $z$ that enhances prompt construction with local, context-aware information. This aligns with the core principle of RAG, which provides ``documents" to transform general models into specialized ones. The retrieved contextual information $c$ serves two purposes in this step: supporting codebase embedding and facilitating content search.

In (\ref{h=usv}), we discussed the use of embeddings in content retrieval, where the embedding functions $\psi(\cdot)$ and $\phi(\cdot)$ project the original sequences to a low-dimensional embedding space for subsequent computation. To balance the modeling power of neural network based encoders, such as BERT \cite{devlin2018bert}, with the computational efficiency of lightweight methods like one-hot embedding, we propose and employ dynamic code symbol indexing (``DCSI"). DCSI enables precise source code analysis by capturing each coding token's symbol information, position, relationships with neighboring tokens, and dependencies within the programming graph. It also supports dynamic updates, adapting to changes such as codebase edits and maintaining synchronization with the local context. While facilitating efficient content search comparison through comprehensive contextual exploitation, DCSI remains computationally efficient. We thus have the following simplified model
\begin{equation} \label{emb_dcsi}
    p_{\eta}(z|x, c) = \frac{\exp{(\mathrm{emb}_\mathrm{DCSI}(z)^TH\mathrm{emb}_\mathrm{DCSI}(x))}}{\sum_{z'}{\exp{(\mathrm{emb}_\mathrm{DCSI}(z')^TH\mathrm{emb}_\mathrm{DCSI}(x))}}}
\end{equation}
where the consistent embedding function makes the heuristic $H$ a square matrix. 

We present a gradient descent algorithm to obtain the optimal values of $H$ and other parameters, as detailed in Section \ref{parameter_tuning}. Given the embedding function and heuristic matrix, the content retriever identifies
\begin{equation*}
    \mathcal{R}_{\eta}(x,c) = \argmax_{z \in \mathrm{emb}(x)} p(c|H, q*)
\end{equation*}
where $q$ represents the optional user query which is provided in cases involving user interactions, such as in Q\&A scenarios.

\begin{figure}[t]
\centering
\includegraphics[width=1.0\linewidth]{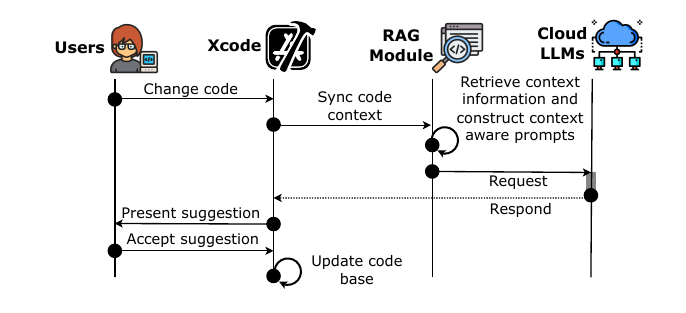}
\caption{Sequence Diagram of \camp{} on {\it Copilot for Xcode} which enables real-time code generation and suggestions.}\label{fig:copilot_for_xcode_fetch_suggestions}
\vspace{-5mm}
\end{figure}
\subsubsection{Prompt Constructor} \label{sec:prompt_construction}
The final component of the RAG module is the prompt constructor $\mathcal{G}(y_i|x, c, z, y_{1:i-1})$, which integrates the retrieved context, content, and the interaction history with the user to form the new prompt.

Table \ref{tab:components} lists the main components included in the constructed prompt, each assigned a priority value for ranking based on experimental trials. When the provided content exceeds the context window limit, lower-priority components are truncated first. The ranking also influences the LLM's performance: for example, when contextual messages are placed after the message history, the LLMs tend to disregard the retrieved information and generate responses primarily based on the message history.

The goal of the prompt constructor is to determine the optimal combination and ranking of the components. Denote the $i$th prompt as $y_i$ and the $k$th configurable component as $y^k$. Without loss of generality, let $\tau_k$ represent the maximum number of configurable components. Each $y_i$ is thus an ordered array of $y^k$. Consequently, we have
\begin{align} \label{eqn_g}
    \mathcal{G}_\theta(x,c,z,y_{1:i-1}) & = y_i \nonumber\\
    & = \mathrm{order}([y^1,y^2,\dots,y^{\tau_k}] ) \\
    & = {\begin{bmatrix}
\theta_1 &
\theta_2 &
\dots &
\theta_k
\end{bmatrix}}^T {\begin{bmatrix}
y^1&
y^2&
\dots&
y^k
\end{bmatrix}}^T \nonumber
\end{align}
where $\theta_k$ are standard unit vectors that mark the component located on the $k$th position of $y_i$. The optimal $\theta$ is determined using Algorithm \ref{alg2}, as described in Section \ref{parameter_tuning}.

\begin{algorithm}
\caption{Train Weight and Heuristic Parameters of Retrievers}\label{alg1}
\begin{algorithmic}
\Require $\Sigma_i\eta'_i = 1, \theta_i$ are standard basis vectors
\Ensure $\tau_H > 0, \tau_{\eta'} > 0, \alpha^n > 0, \beta^n > 0$
\State $H^1 = H^0 \in \mathbb{R}^{d_{\mathrm{emb}} \times d_{\mathrm{emb}}}, \eta'^0 = \eta'^1 \in \mathbb{R}^{d_c}$
\State $c_0 \in \mathbb{R}^{d_c}, z_0 \in \mathbb{R}^{d_z}$
\State $t^0 \gets t^1 \gets 1, n \gets 1$
\While{not converged}
\State $\bar{H}^n \gets H^n + \frac{t^{n-1}-1}{t^n}(H^n-H^{n-1})$
\State $G^n \gets \bar{H}^n - \frac{1}{\tau_H}\nabla_{\bar{H}^n}(\mathcal{L}(\mathcal{X},\mathcal{Y},H))$
\State $[U \Sigma V] \gets \mathrm{SVD}(G^n)$
\State $\bar{H}^n \gets U\mathcal{D}_{\tau_H}(\Sigma)V^T$ \Comment{$D_\tau{X}=\max(X-\tau, 0)$}
\State $H^{n+1} \gets H^n + \alpha^n(\bar{H}^n - h^n)$
\State $c = \mathcal{R}_{\eta\supset{H^{n+1}}}^{-1}(z)$
\State $\bar{\eta}'^n \gets \eta'^n + \frac{t^{n-1}-1}{t^n}(\eta'^n-\eta'^{n-1})$
\State $g^n \gets \bar{\eta}'^n - \frac{1}{\tau_{\eta'}}\nabla_{\bar{\eta}'^n}(\mathcal{L}(\mathcal{X},\mathcal{Y},\eta'))$
\State $\eta'^{n+1} \gets \eta'^n + \beta^n(\bar{\eta}'^n - \eta'^n)$
\State $t^{n+1} \gets \frac{1+\sqrt{1+4(t^n)^2}}{2}$
\State $n \gets n + 1$
\EndWhile
\end{algorithmic}
\end{algorithm}
\begin{theorem}
Starting from any initial weight ($\eta'$) and heuristic ($H$) parameters of the retrievers, the optimal solution can be obtained iteratively by Algorithm \ref{alg1}.
\end{theorem}

\subsection{Parameter Tuning} \label{parameter_tuning}
This section presents the algorithms for model parameter tuning, including: 1) a gradient descent-based algorithm for computing the weight parameter $\eta'$ and heuristic matrix $H$ and 2) a sorting algorithm for computing the ranking parameter $\theta$.

\begin{algorithm}
\caption{Train Ranking Parameters of Prompt Constructor}\label{alg2}
\begin{algorithmic}
\State initialize directional graph $G$
\For{\texttt{all possible} $(\theta_i, \theta_j)$} 
    \If{$\mathcal{L} - \mathcal{L}_{i \xleftrightarrow{} j} > \epsilon$}
    \State \texttt{store directional edge $i-j$ to $G$}
\EndIf
\EndFor
\State run topological sort on $G$ \\
\Return $G$
\end{algorithmic}
\end{algorithm}
\begin{theorem}
The optimal ordering of the $k$ prompt components $y^{1:k}$ can be solved by Algorithm \ref{alg2} in $\mathcal{O}(k^2)$.
\end{theorem}

To solve the optimization problem presented by (\ref{opt1}), we introduce Algorithm \ref{alg1} to train the weight and heuristic parameters of the retrievers iteratively, by alternatively moving $H$ and $\eta'$ to the negative gradient direction in each iteration $n$, with step size $\alpha$ and $\beta$ correspondingly.

Given trained context and content retrievers, we obtain the individual prompt components, $y^{1:k}$, as deterministic outcomes from the retrievers. To determine the optimal ranking parameter $\theta$, as described in (\ref{eqn_g}), we present Algorithm \ref{alg2}. A brute-force approach would involve traversing all possible arrangements of the $k$ components, yielding a time complexity of $\mathcal{O}(k!)$. However, we observe that, in most cases, rearranging a subset of the $k$ components leads to only trivial differences in performance. Significant improvements typically arise when swapping just two components. As a result, we model the $k$ components as nodes in a directed graph, where an edge represents the topological relationship between neighboring components. In Algorithm \ref{alg2}, we first test the topological relationship between all pairs by switching them and comparing the outcomes. We then apply a topological sort to determine a reasonable order, reducing the time complexity to $\mathcal{O}(k^2 + k + C) \rightarrow \mathcal{O}(k^2)$, where $C$ is the number of edges, which is considered a small constant in our use cases.

\begin{table*}[h]
    \centering
    \caption{Evaluation Results for Code Generation Tasks on CoderEval. The performance of \camp{} is compared to baseline models.}
    \begin{tabular}{l|ccc|ccc|ccc}
        \toprule
        \textbf{Model} & \multicolumn{3}{c|}{\textbf{class-runnable}} & \multicolumn{3}{c|}{\textbf{file-runnable}} & \multicolumn{3}{c}{\textbf{project-runnable}} \\
         & Pass@1 & Pass@5 & Pass@10 & Pass@1 & Pass@5 & Pass@10 & Pass@1 & Pass@5 & Pass@10 \\
        \midrule
        CloudOnly & 8.73\% & 12.57\% & 14.55\% & 21.03\% & 29.09\% & 32.35\% & 9.37\% & 12.08\%  & 13.04\% \\
        BaseRAG & 19.84\% & 35.06\% & 40.91\% & 24.98\% & 35.94\% & 39.01\% & 15.66\% & 21.89\% & 24.62\% \\
        FileContext & \textbf{31.23\%} & \textbf{43.41\%} & \textbf{47.30\%} & 29.52\% & 37.80\% & 42.30\% & 11.08\% & 16.87\% & 17.92\% \\
        \camp{} & 28.96\% & 41.72\% & 46.07\% & \textbf{35.30\%} & \textbf{43.45\%} & \textbf{45.80\%} & \textbf{21.91\%} & \textbf{25.05\%} & \textbf{26.43\%} \\
        \bottomrule
    \end{tabular}
    \label{tab:obj}
\end{table*}

\subsection{Implementation Details on Xcode}
We demonstrate the practical utility of \camp{} by implementing it as a plugin for Xcode. This serves as a pilot trial to validate the methodology's robustness in challenging coding environments with sandboxed architecture that imposes strict restrictions and offers limited access to local contextual information. To address these challenges, we employed: 1) XPC service-level communication to enable interaction with language servers and facilitate real-time code suggestions in the UI, and 2) the {\tt Accessibility API} to capture rich contextual data. These solutions enable accurate prompt construction and effective integration with the IDE environment, laying the groundwork for future expansions to other IDEs.

The system workflow is illustrated in Figure~\ref{fig:copilot_for_xcode_fetch_suggestions}. When users update their code, \camp{} retrieves contextual information, constructs enriched prompts, and facilitates real-time AI-assisted programming. The system dynamically interacts with Xcode to deliver tailored code suggestions and handle questions through the chat panel, thereby enhancing developer productivity and overall coding experience.

\section{Evaluation}
We evaluate the performance of \camp{} using code generation benchmarks and user studies. The results demonstrate its superiority over baseline models in code completion tasks across varying complexities, as well as its effectiveness in real-world programming scenarios.
\subsection{Experiment Setup}
\subsubsection{Dataset and Evaluation Metrics}
We employ the CoderEval benchmark \cite{coder_eval}, a pragmatic code generation evaluation dataset designed to measure the performance of generative pre-trained models. Compare to benchmarkes like HumanEval \cite{chen2021evaluating} which focuses on standalone functions, CoderEval includes cross-class and cross-file test cases, effectively evaluating model performance on larger projects and repositories. The benchmark comprises 230 test cases categorized into six runnable levels, from single-function to project-level tasks. For our experiments, we selected the top three categories with the highest runnable levels, representing the most common real-world use cases and encompassing diverse contexts:
\begin{itemize}
    \item \textit{class-runnable}: Code outside the function but within the same class.
    \item \textit{file-runnable}: Code outside the class but within the same file. 
     \item \textit{project-runnable}: Code outside the file, spanning multiple files or repositories.
\end{itemize}

The performance is measured by Pass@K, defined as\cite{chen2021evaluating}:
\begin{equation*}
    \mathrm{Pass@K} = \mathbb{E}\left(1 - \frac{\binom{n-c}{k}}{\binom{n}{k}}\right),
\end{equation*}
where $n$ is the total number of samples, $c$ is the number of correct samples, and $k$ is the number of top-generated solutions considered.

\subsubsection{Baseline Models}
We compare \camp{} against the following baseline models, using GPT-3.5-Turbo as the cloud-based LLM. 
\begin{itemize}
    \item \textit{CloudOnly}: Inputs are processed solely by the cloud-based model, with no local processing or context retrieval.
    \item \textit{BaseRAG}: Implements standard RAG techniques as proposed by \cite{rag_lewis}.
    \item \textit{FileContext}: A variant of \camp{} that prioritizes context retrieved from the currently open files in the IDE. This lightweight version balances performance and resource efficiency.
\end{itemize}

\subsection{Results and Analysis}
The objective evaluation results, summarized in Table \ref{tab:obj}, show that \camp{} consistently outperforms the baseline models across all runnable levels, with more significant performance improvements at higher levels. 

Typically, it achieves a 12.5\% and 6.3\% improvement over CloudOnly and BaseRAG, respectively, in Pass@1 accuracy for the project-runnable category. Compared to the CloudOnly model, \camp{} achieves advantageous results in all tasks, demonstrating the impact of retrieved content in enhancing LLM prompts. Similarly, \camp{} outperforms the BaseRAG model, highlighting the effectiveness of its context-based retrieval mechanisms in understanding the codebase and generating context-aware solutions. 

The FileContext model shows comparable performance to \camp{} for lower runnable levels, such as class-runnable tasks, but falls behind in cross-file and project-level scenarios. This outcome emphasizes the necessity of broader context retrieval, a key advantage enabled by RAG techniques. The results also suggest that dynamically adjusting the retrieval scope based on task complexity can optimize computational resource without compromising accuracy. For instance, narrowing the retrieval range to specific files for class-level tasks can reduce computational overhead while maintaining high performance.

\subsection{User Studies}
To further evaluate the practicality of \camp{}, we conducted user studies involving 14 iOS developers. Participants were tasked with completing six programming assignments selected from the Software-artifact Infrastructure Repository: three focused on code completion and error debugging, two on database operations, and one on UI creation. The test group used {\it Copilot for Xcode}, while the control group worked without it. Completion times were recorded for comparison.

The results show that the test group achieved a 37.2\% reduction in completion time compared to the control group, with a 45\% code suggestion adoption rate. Qualitative observations reveal that context-aware code generation provided notable advantages. For instance, in error debugging tasks involving nested class dependencies, {\it Copilot for Xcode} efficiently generated correct fixes using cross-repository context; for the UI creation task, the tool generated a boilerplate home view aligned with similar pages in the repository, thus significantly reducing manual effort.

Conversely, areas for improvement were identified, including ``cold-start" issues, where generation slowed immediately after loading large repositories or during bulk edits involving extensive codebase indexing. These findings underscore opportunities for further optimization of \camp{} and its tooling implementation.

\section{Conclusion}
This paper presented \camp{}, a multi-model programming copilot solution that leverages context-based RAG to enhance AI-assisted programming. By introducing dynamic context retrieval from local codebases, \camp{} optimizes context-aware prompt construction, bridging the gap between the generative capabilities of cloud-based LLMs and the contextual efficiency of local models. It also fosters dynamic collaboration between cloud LLMs and local models, paving the way for advanced AI-assisted programming solutions. As a further extension to AI-assisted code generation, our latest research extends to AI-assisted software testing, where LLMs facilitate automated test generation, bug detection, and secure code refinement \cite{copilot_for_testing}. By enabling seamless integration of human expertise with AI tools, \camp{} aligns with Dijkstra's vision of augmenting human intelligence in software development, advancing toward more efficient, reliable, and user-centric programming practices.

\section*{Acknowledgment}
This research was supported by the Singapore Ministry of Education AcRF under Grant RG91/22 and NTU startup fund.

\bibliographystyle{unsrtnat}
\bibliography{references}  

\begin{thebibliography}{40}
\providecommand{\natexlab}[1]{#1}
\providecommand{\url}[1]{\texttt{#1}}
\expandafter\ifx\csname urlstyle\endcsname\relax
  \providecommand{\doi}[1]{doi: #1}\else
  \providecommand{\doi}{doi: \begingroup \urlstyle{rm}\Url}\fi

\bibitem[Dijkstra((transcribed) 2007)]{dijkstra}
Edsger~Wybe Dijkstra.
\newblock A preliminary investigation into computer assisted programming.
\newblock \emph{E. W. Dijkstra Archive (EWD 237)}, (transcribed) 2007.

\bibitem[Dijkstra(1972)]{dijkstra1972humble}
Edsger~Wybe Dijkstra.
\newblock The humble programmer.
\newblock \emph{Communications of the ACM}, 15\penalty0 (10):\penalty0 859--866, 1972.

\bibitem[Mozannar et~al.(2022)Mozannar, Bansal, Fourney, and Horvitz]{mozannar2022reading}
Hussein Mozannar, Gagan Bansal, Adam Fourney, and Eric Horvitz.
\newblock Reading between the lines: Modeling user behavior and costs in {AI}-assisted programming.
\newblock \emph{preprint arXiv:2210.14306}, 2022.

\bibitem[Wong et~al.(2023)Wong, Guo, Hang, Ho, and Tan]{wong2023natural}
Man-Fai Wong, Shangxin Guo, Ching-Nam Hang, Siu-Wai Ho, and Chee-Wei Tan.
\newblock Natural language generation and understanding of big code for {AI}-assisted programming: A review.
\newblock \emph{Entropy}, 25\penalty0 (6):\penalty0 888, 2023.

\bibitem[Sammet(1966)]{Sammet1966}
Jean~E. Sammet.
\newblock The use of {English} as a programming language.
\newblock \emph{Communications of the {ACM}}, 9\penalty0 (3):\penalty0 228--230, mar 1966.
\newblock \doi{10.1145/365230.365274}.
\newblock URL \url{https://doi.org/10.1145/365230.365274}.

\bibitem[Waters(1982)]{MITapprentice82}
Richard~C. Waters.
\newblock The programmer's apprentice: Knowledge based program editing.
\newblock \emph{IEEE Transactions on Software Engineering}, SE-8\penalty0 (1):\penalty0 1--12, January 1982.
\newblock \doi{10.1109/TSE.1982.234769}.

\bibitem[Rich and Waters(1988)]{MITapprentice88}
Charles Rich and Richard~C. Waters.
\newblock The programmer's apprentice: a research overview.
\newblock \emph{Computer}, 21\penalty0 (11):\penalty0 10--25, November 1988.
\newblock \doi{10.1109/2.86782}.

\bibitem[Handsaker(1982)]{MITcodegeneration}
Robert~E. Handsaker.
\newblock Code generation in the programmer's apprentice.
\newblock Working Paper 233, {MIT} {AI} Lab, May 1982.

\bibitem[Rich et~al.(1978)Rich, Shrobe, Waters, Sussman, and Hewitt]{MITprompt}
Charles Rich, Howard~E. Shrobe, Robert~C. Waters, Gerald~J. Sussman, and Carl~E. Hewitt.
\newblock Programming viewed as an engineering activity.
\newblock {AI} Memo 459, Massachusetts Institute of Technology, January 1978.

\bibitem[Rich and Waters(1982)]{MITsimplify}
Charles Rich and Richard~C. Waters.
\newblock The disciplined use of simplifying assumptions.
\newblock \emph{ACM SIGSOFT Software Engineering Notes}, 7\penalty0 (5):\penalty0 150--154, December 1982.

\bibitem[Codeium(2023)]{codeium}
Exafunction Codeium.
\newblock Codeium - free {AI} code completion \& chat.
\newblock \url{https://codeium.com/}, 2023.
\newblock Accessed on June 1, 2023.

\bibitem[Git()]{GitHubCopilot}
{GitHub Copilot · Your {AI} pair programmer}.
\newblock Online.
\newblock URL \url{https://copilot.github.com/}.
\newblock Accessed: Mar. 19, 2025.

\bibitem[Pearce et~al.(2025)Pearce, Ahmad, Tan, Dolan-Gavitt, and Karri]{GitHubCopilot_2022}
Hammond Pearce, Baleegh Ahmad, Benjamin Tan, Brendan Dolan-Gavitt, and Ramesh Karri.
\newblock Asleep at the keyboard? assessing the security of {Github Copilot}’s code contributions.
\newblock \emph{Communications of the ACM}, 68\penalty0 (2):\penalty0 96--105, 2025.

\bibitem[OpenAI(2023)]{gpt42023}
OpenAI.
\newblock Gpt-4 technical report.
\newblock \emph{preprint arXiv:2303.08774}, 2023.

\bibitem[Amazon(2022)]{amazon}
CodeWhisperer Amazon.
\newblock {AI} code generator - amazon codewhisperer.
\newblock \url{https://aws.amazon.com/codewhisperer}, 2022.
\newblock Accessed on June 1, 2023.

\bibitem[Roziere et~al.(2023)Roziere, Gehring, Gloeckle, Sootla, Gat, Tan, Adi, Liu, Sauvestre, Remez, et~al.]{meta_llama}
Baptiste Roziere, Jonas Gehring, Fabian Gloeckle, Sten Sootla, Itai Gat, Xiaoqing~Ellen Tan, Yossi Adi, Jingyu Liu, Romain Sauvestre, Tal Remez, et~al.
\newblock Code llama: Open foundation models for code.
\newblock \emph{preprint arXiv:2308.12950}, 2023.

\bibitem[Zheng et~al.(2015)Zheng, Joe-Wong, Tan, Chiang, and Wang]{zheng2015bid}
Liang Zheng, Carlee Joe-Wong, Chee~Wei Tan, Mung Chiang, and Xinyu Wang.
\newblock How to bid the cloud.
\newblock In \emph{Proceedings of the 2015 ACM Conference on Special Interest Group on Data Communication (SIGCOMM)}, pages 71--84, 2015.

\bibitem[Lewis et~al.(2020)Lewis, Perez, Piktus, Petroni, Karpukhin, Goyal, K{\"u}ttler, Lewis, Yih, Rockt{\"a}schel, et~al.]{rag_lewis}
Patrick Lewis, Ethan Perez, Aleksandra Piktus, Fabio Petroni, Vladimir Karpukhin, Naman Goyal, Heinrich K{\"u}ttler, Mike Lewis, Wen-tau Yih, Tim Rockt{\"a}schel, et~al.
\newblock Retrieval-augmented generation for knowledge-intensive {NLP} tasks.
\newblock \emph{Advances in Neural Information Processing Systems}, 33:\penalty0 9459--9474, 2020.

\bibitem[Tan et~al.(2023)Tan, Guo, Wong, and Hang]{tan2023copilot}
Chee~Wei Tan, Shangxin Guo, Man~Fai Wong, and Ching~Nam Hang.
\newblock {Copilot for Xcode}: exploring {AI}-assisted programming by prompting cloud-based large language models.
\newblock \emph{preprint arXiv:2307.14349}, 2023.

\bibitem[GitHub(2024)]{github_xcode_2024}
GitHub.
\newblock Github copilot code completion in xcode is now available in public preview.
\newblock \url{https://github.blog/changelog/2024-10-29-github-copilot-code-completion-in-xcode-is-now-available-in-public-preview/}, October 2024.
\newblock Accessed: January 5, 2025.

\bibitem[Hindle et~al.(2012)Hindle, Barr, Su, Gabel, and Devanbu]{hindle2012naturalness}
Abram Hindle, Earl~T Barr, Zhendong Su, Mark Gabel, and Premkumar Devanbu.
\newblock On the naturalness of software.
\newblock In \emph{2012 34th International Conference on Software Engineering}, pages 837--847. IEEE, 2012.

\bibitem[Vaswani et~al.(2017)Vaswani, Shazeer, Parmar, Uszkoreit, Jones, Gomez, Kaiser, and Polosukhin]{vaswani2017attention}
Ashish Vaswani, Noam Shazeer, Niki Parmar, Jakob Uszkoreit, Llion Jones, Aidan~N Gomez, {\L}ukasz Kaiser, and Illia Polosukhin.
\newblock Attention is all you need.
\newblock \emph{Advances in Neural Information Processing Systems}, 2017.

\bibitem[Vechev et~al.(2016)Vechev, Yahav, et~al.]{vechev2016programming}
Martin Vechev, Eran Yahav, et~al.
\newblock Programming with “big code”.
\newblock \emph{Foundations and Trends{\textregistered} in Programming Languages}, 3\penalty0 (4):\penalty0 231--284, 2016.

\bibitem[Raffel et~al.(2020)Raffel, Shazeer, Roberts, Lee, Narang, Matena, Zhou, Li, and Liu]{raffel2020exploring}
Colin Raffel, Noam Shazeer, Adam Roberts, Katherine Lee, Sharan Narang, Michael Matena, Yanqi Zhou, Wei Li, and Peter~J Liu.
\newblock Exploring the limits of transfer learning with a unified text-to-text transformer.
\newblock \emph{The Journal of Machine Learning Research}, 2020.

\bibitem[Devlin et~al.(2018)Devlin, Chang, Lee, and Toutanova]{devlin2018bert}
Jacob Devlin, Ming-Wei Chang, Kenton Lee, and Kristina Toutanova.
\newblock Bert: Pre-training of deep bidirectional transformers for language understanding.
\newblock \emph{preprint arXiv:1810.04805}, 2018.

\bibitem[Anil et~al.(2023)Anil, Dai, Firat, Johnson, Lepikhin, Passos, Shakeri, Taropa, Bailey, Chen, et~al.]{anil2023palm}
Rohan Anil, Andrew~M Dai, Orhan Firat, Melvin Johnson, Dmitry Lepikhin, Alexandre Passos, Siamak Shakeri, Emanuel Taropa, Paige Bailey, Zhifeng Chen, et~al.
\newblock Palm 2 technical report.
\newblock \emph{preprint arXiv:2305.10403}, 2023.

\bibitem[Waldinger and Lee(1969)]{waldinger1969prow}
Richard~J Waldinger and Richard~CT Lee.
\newblock Prow: A step toward automatic program writing.
\newblock In \emph{1st International Joint Conference on Artificial Intelligence}, pages 241--252, 1969.

\bibitem[Wong and Tan(2024)]{wong2024aligning}
Man~Fai Wong and Chee~Wei Tan.
\newblock Aligning crowd-sourced human feedback for reinforcement learning on code generation by large language models.
\newblock \emph{IEEE Transactions on Big Data}, 2024.

\bibitem[Robbes and Lanza(2008)]{robbes2008program}
Romain Robbes and Michele Lanza.
\newblock How program history can improve code completion.
\newblock In \emph{23rd IEEE/ACM International Conference on Automated Software Engineering}, pages 317--326, 2008.

\bibitem[Acharya et~al.(2007)Acharya, Xie, Pei, and Xu]{acharya2007mining}
Mithun Acharya, Tao Xie, Jian Pei, and Jun Xu.
\newblock Mining {API} patterns as partial orders from source code: From usage scenarios to specifications.
\newblock In \emph{6th Joint Meeting of The European Software Engineering Conference and The ACM SIGSOFT Symposium on The Foundations of Software Engineering}, pages 25--34, 2007.

\bibitem[Wang et~al.(2025)Wang, Guo, and Tan]{copilot_for_testing}
Yuchen Wang, Shangxin Guo, and Chee~Wei Tan.
\newblock From code generation to software testing: {AI Copilot} with context-based {RAG}.
\newblock \emph{IEEE Software}, pages 1--9, 2025.
\newblock \doi{10.1109/MS.2025.3549628}.

\bibitem[Izacard and Grave(2020)]{rag_izacard}
Gautier Izacard and Edouard Grave.
\newblock Leveraging passage retrieval with generative models for open domain question answering.
\newblock \emph{preprint arXiv:2007.01282}, 2020.

\bibitem[{Apple Inc.}(2023)]{xcode}
{Apple Inc.}
\newblock \emph{{Xcode 15}}, 2023.
\newblock URL \url{https://developer.apple.com/xcode/}.
\newblock Accessed: Mar. 19, 2025.

\bibitem[Hellendoorn et~al.()Hellendoorn, Devanbu, and Bacchelli]{ide_8}
Vincent~J Hellendoorn, Premkumar~T Devanbu, and Alberto Bacchelli.
\newblock Will they like this? evaluating code contributions with language models.
\newblock In \emph{2015 IEEE/ACM 12th Working Conference on Mining Software Repositories}.

\bibitem[Feng et~al.(2020)Feng, Guo, Tang, Duan, Feng, Gong, Shou, Qin, Liu, Jiang, et~al.]{ide_9}
Zhangyin Feng, Daya Guo, Duyu Tang, Nan Duan, Xiaocheng Feng, Ming Gong, Linjun Shou, Bing Qin, Ting Liu, Daxin Jiang, et~al.
\newblock Codebert: A pre-trained model for programming and natural languages.
\newblock \emph{preprint arXiv:2002.08155}, 2020.

\bibitem[{Apple Inc.}(2021)]{ide_10_apple}
{Apple Inc.}
\newblock App sandbox, 2021.
\newblock URL \url{https://developer.apple.com/library/archive/documentation/Security/Conceptual/AppSandboxDesignGuide/AboutAppSandbox/AboutAppSandbox.html}.
\newblock Accessed: \today.

\bibitem[Rosenfeld et~al.(1996)]{max_entropy}
Ronald Rosenfeld et~al.
\newblock A maximum entropy approach to adaptive statistical language modelling.
\newblock \emph{Computer speech and language}, 10\penalty0 (3):\penalty0 187, 1996.

\bibitem[Toh and Yun(2010)]{proximal_gradient_algo}
Kim-Chuan Toh and Sangwoon Yun.
\newblock An accelerated proximal gradient algorithm for nuclear norm regularized linear least squares problems.
\newblock \emph{Pacific Journal of optimization}, 6\penalty0 (615-640):\penalty0 15, 2010.

\bibitem[Yu et~al.(2024)Yu, Shen, Ran, Zhang, Zhang, Ma, Liang, Li, Wang, and Xie]{coder_eval}
Hao Yu, Bo~Shen, Dezhi Ran, Jiaxin Zhang, Qi~Zhang, Yuchi Ma, Guangtai Liang, Ying Li, Qianxiang Wang, and Tao Xie.
\newblock Codereval: A benchmark of pragmatic code generation with generative pre-trained models.
\newblock In \emph{Proceedings of the 46th IEEE/ACM International Conference on Software Engineering}, pages 1--12, 2024.

\bibitem[Chen et~al.(2021)Chen, Tworek, Jun, Yuan, Pinto, Kaplan, Edwards, Burda, Joseph, Brockman, et~al.]{chen2021evaluating}
Mark Chen, Jerry Tworek, Heewoo Jun, Qiming Yuan, Henrique Ponde de~Oliveira Pinto, Jared Kaplan, Harri Edwards, Yuri Burda, Nicholas Joseph, Greg Brockman, et~al.
\newblock Evaluating large language models trained on code.
\newblock \emph{preprint arXiv:2107.03374}, 2021.

\end{thebibliography}






\end{document}